\documentclass{article}

\usepackage{microtype}
\usepackage{graphicx}
\usepackage{subcaption}
\usepackage{booktabs}
\usepackage{multirow}

\usepackage{hyperref}

\usepackage[preprint]{icml2026}

\usepackage{amsmath}
\usepackage{amssymb}
\usepackage{mathtools}
\usepackage{amsthm}

\usepackage[capitalize,noabbrev]{cleveref}

\theoremstyle{plain}

\theoremstyle{definition}

\theoremstyle{remark}

\usepackage[textsize=tiny]{todonotes}

\icmltitlerunning{}

\begin{document}

\twocolumn[
  \icmltitle{ConfLayers: Adaptive Confidence-based Layer Skipping \\for Self-Speculative Decoding}
  \icmlsetsymbol{equal}{*}

  \begin{icmlauthorlist}
    \icmlauthor{Walaa Amer}{yyy,comp}
    \icmlauthor{Uday Das}{comp}
    \icmlauthor{Fadi Kurdahi}{yyy}
  \end{icmlauthorlist}

  \icmlaffiliation{yyy}{Department of Electrical Engineering and Computer Sciences, University of California Irvine, Irvine, USA}
  \icmlaffiliation{comp}{Advanced Micro Devices, San Jose, USA}

  \icmlcorrespondingauthor{Walaa Amer}{walaaa@uci.edu}

  \icmlkeywords{}

  \vskip 0.3in
]

\printAffiliationsAndNotice{}  
\begin{abstract}
Self-speculative decoding is an inference technique for large language models designed to speed up generation without sacrificing output quality. It combines fast, approximate decoding using a compact version of the model as a draft model with selective re-evaluation by the full target model. Some existing methods form the draft model by dynamically learning which layers to skip during inference, effectively creating a smaller subnetwork to speed up computation. However, using heuristic-based approaches to select layers to skip can often be simpler and more effective.
In this paper, we propose ConfLayers, a dynamic plug-and-play approach to forming the draft model in self-speculative decoding via confidence-based intermediate layer skipping. The process iteratively computes confidence scores for all layers, selects layers to skip based on an adaptive threshold, evaluates the performance of the resulting set, and updates the best selection until no further improvement is achieved or a maximum number of iterations is reached.
This framework avoids the overhead and complexity of training a layer skipping policy and can provide more consistent speed-quality trade-offs while preserving the adaptivity of the draft model to diverse tasks and datasets. 
The performance evaluation of ConfLayers across different models and datasets shows that our novel approach offers up to 1.4$\times$ speedup over vanilla LLM generation. The code is available on \url{https://github.com/WA225/ConfLayers}.
\end{abstract}

\section{Introduction}

The increasing scale and capability of large language models (LLMs) have transformed natural language processing, enabling strong performance across tasks such as question answering, summarization, machine translation, and reasoning. Modern LLMs, often comprising tens or hundreds of billions of parameters, exhibit strong generalization and emergent reasoning abilities, but these advances come at a substantial computational and latency cost. Due to the autoregressive nature of inference, decoding remains particularly expensive. As model sizes and deployment scales continue to grow, achieving \textbf{real-time, resource-efficient inference} has become a key bottleneck for deploying LLMs in interactive applications.

\begin{figure}[t]
\centering
\includegraphics[width=.99\columnwidth]{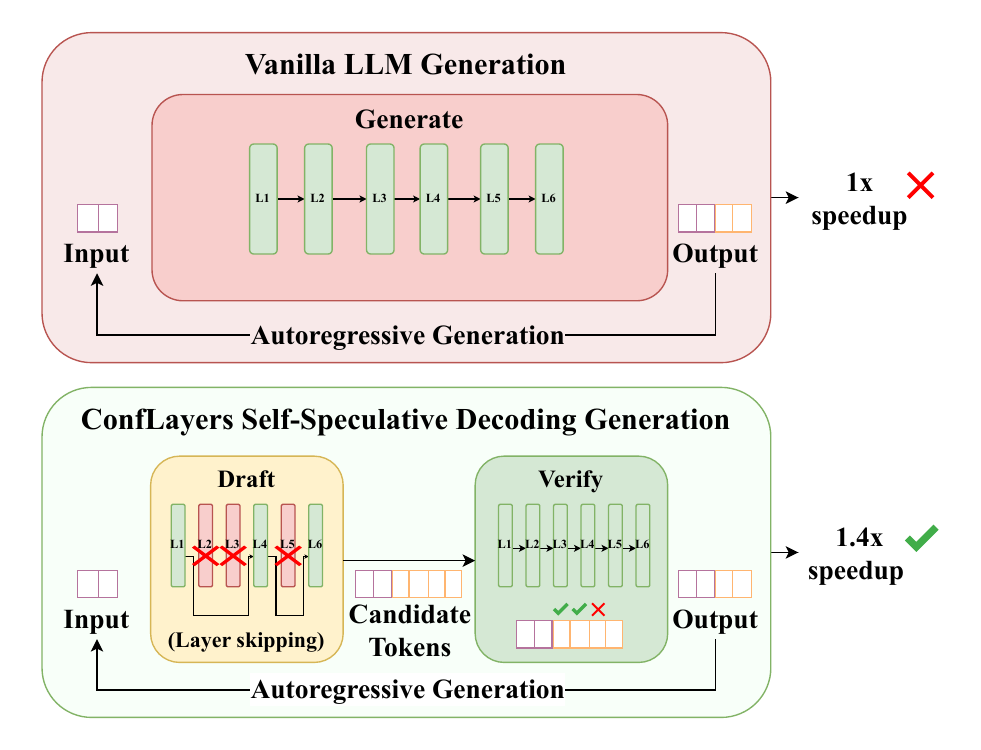}
\caption{\centering ConfLayers, a self-speculative decoding framework with layer skipping, offers up to 1.4$\times$ speedup compared to vanilla LLM generation.}
\label{fig11}
\end{figure}

A wide range of approaches has been explored to reduce inference cost. Techniques such as model compression, quantization, and early-exit inference aim to lower per-token computation but do not address the fundamental sequential dependency of autoregressive decoding. In contrast, \textbf{speculative decoding} has emerged as a promising paradigm for alleviating this bottleneck. In speculative decoding, a draft model rapidly generates multiple tokens ahead, which are then verified by the target model in a single forward pass. This process enables substantial inference acceleration while preserving output quality.

While effective, traditional speculative decoding introduces several practical challenges. Using a separate draft model requires additional training, parameter storage, and synchronization, complicating deployment and scaling. Distribution mismatches between the draft and target models can further reduce token acceptance rates and limit speedup.

To address these limitations, \textbf{self-speculative decoding} (SSD) reuses the same LLM as both the draft and verifier (Figure 1), eliminating the need for an auxiliary model. This is typically achieved by constructing a lightweight subnetwork for speculative generation, while the full model verifies the output. By reusing parameters and internal representations, SSD preserves the benefits of speculative decoding while avoiding external model overhead, motivating \textbf{training-free, plug-and-play} draft construction.

A central challenge in SSD is designing efficient and reliable draft subnetworks. Effective drafts must reduce computation while maintaining sufficient representational fidelity for high acceptance rates. Prior approaches rely on policy-based or heuristic layer-skipping strategies: policy-based methods incur high training cost and limited generalization, while heuristic rules are simpler but lack adaptivity, resulting in suboptimal speed–quality trade-offs.

We introduce \textbf{ConfLayers}, a confidence-driven and training-free framework for constructing adaptive draft subnetworks in self-speculative decoding (Figure 1). ConfLayers estimates intermediate-layer confidence during generation to assess each layer’s contribution to predictive certainty, selectively skipping layers with low confidence impact. The draft configuration is iteratively refined based on verifier acceptance, producing a compact and context-aware subnetwork without retraining or architectural modification.

Across diverse benchmarks and multiple models, ConfLayers achieves up to \textbf{1.4× speedup} over standard autoregressive decoding while maintaining high output quality. It consistently outperforms prior heuristic and dynamic skipping or early-exiting baselines such as SWIFT and DEL, demonstrating robust and efficient acceleration.

Our main contributions are:

\begin{itemize}
    \item We propose a novel \textbf{confidence-based} mechanism for constructing draft subnetworks in self-speculative decoding, eliminating the need for auxiliary models or training.
    \item We design an iterative \textbf{plug-and-play adaptation algorithm} that balances decoding speed and token acceptance rate through confidence-guided layer selection.
    \item We conduct a \textbf{comprehensive empirical evaluation} across multiple datasets and LLM architectures, showing consistent acceleration and quality retention compared to existing baselines.
\end{itemize}

ConfLayers provides a scalable and generalizable solution for accelerating LLM decoding, highlighting the promise of confidence-guided inference for efficient LLM generation.

\begin{figure*}[t]
    \centering
    \includegraphics[width=\textwidth]{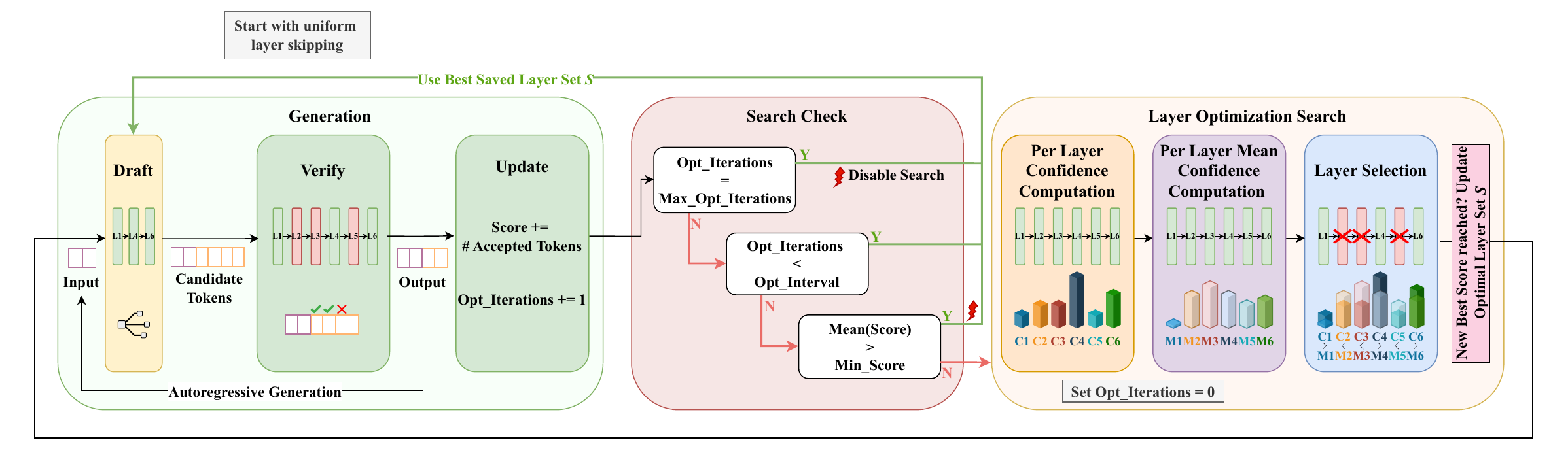}
    \caption{\centering ConfLayers framework implements speculative decoding in a 3-step process. The generation step will use the current non-skip layer set to form the draft model and generated candidate tokens that will be verified via the target model. The layer optimization search step relies on confidence-based layer skipping to form the non-skip layer set. The search check step connects the two previous steps by ensuring the search process is organized and ran at a need-basis.}
    \label{fig6}
\end{figure*}

\section{Related Works}
\label{related}

Speculative decoding has emerged as an effective paradigm for accelerating autoregressive large language model inference by generating multiple candidate tokens in parallel using a lightweight draft model and verifying them with the target model. Early work by Leviathan et al.~\cite{leviathan2023specdec} and Chen et al.~\cite{chen2023specdec} showed that speculative decoding can achieve significant speedups while preserving distributional equivalence to standard decoding. Subsequent extensions explored richer drafting strategies, including feature-space speculation in EAGLE~\cite{li2024eagle}, which reduces uncertainty by extrapolating intermediate hidden states, and tree-based speculation in SEQUOIA~\cite{chen2024sequoia}, which constructs optimized token trees to improve acceptance rates under larger speculation budgets. Other works relax standard assumptions in speculative decoding, such as Lossless SD for Heterogeneous Vocabularies~\cite{timor2025lossless}, which removes the shared-tokenizer requirement between draft and target models, and Reward-Guided Speculative Decoding~\cite{liao2025rewardguided}, which biases drafting toward high-reward continuations to optimize the speed–quality trade-off.

Several approaches further enhance speculative decoding by incorporating external information or parallel generation signals. Nearest Neighbor Speculative Decoding (NEST)~\cite{li2024nearest} augments drafting with retrieval from a text datastore, enabling multi-token speculation grounded in real data and improving robustness on knowledge-intensive tasks. Speculative decoding has also been extended to multi-sample inference settings, where overlapping predictions across parallel generations are used as high-confidence draft tokens~\cite{li2025multisample}. More recently, Saguaro~\cite{anonymous2026speculative} proposes an asynchronous formulation in which draft speculation continues in parallel with verification, further mitigating the sequential bottleneck.

While effective, most speculative decoding methods rely on a separate draft model, increasing memory footprint, training cost, and deployment complexity. To address this limitation, \textbf{self-speculative decoding} (SSD) eliminates auxiliary models by leveraging the internal structure of the original LLM. Instead of an external draft, shallow executions of the same network generate speculative tokens that are verified by deeper layers. Draft \& Verify~\cite{zhang2024draftverify} is the seminal work in this direction, showing that executing only a subset of transformer layers suffices to produce high-quality draft tokens without retraining or architectural modification, achieving up to $1.99\times$ speedup on LLaMA-2 models. Kangaroo~\cite{liu2024kangaroo} extends this idea using a fixed shallow prefix combined with confidence-based early stopping, further reducing wasted computation on uncertain tokens. LayerSkip~\cite{elhoushi2024layerskip} integrates training-time layer dropout with inference-time early exit, enabling intermediate layers to act as valid draft models.

More recent work emphasizes \textbf{adaptive and plug-and-play} SSD strategies. Draft on the Fly~\cite{metel2024draftonthefly} dynamically selects skip layers based on cosine similarity between hidden states. SWIFT~\cite{xia2025swift} further advances this direction by adaptively selecting skip layers at runtime on a per-input basis without additional training, achieving consistent speedups across both open-ended and structured tasks. CLaSP~\cite{chen2025clasp} formulates layer skipping as a dynamic programming problem, updating skip schedules after each verification step to maximize acceptance rates. DEL~\cite{zarch2025del} introduces a heuristic controller that jointly adjusts exit depth and speculative length online, while KNN-SSD~\cite{song2025knnssd} improves robustness under domain shift by retrieving effective skip strategies from a nearest-neighbor memory built over prior inputs.

Overall, many approaches have been proposed for  self-speculative decoding layer-based drafting, ranging from static mechanisms to dynamic and input-adaptive strategies. While these methods perform layer skipping, their decisions operate at coarse levels (skip sets, exit layers…), introducing a significant overhead. In contrast, ConfLayers computes confidence online and enables \textbf{confidence-guided per-layer execution decisions} within the same step, providing quick and \textbf{finer-grained input adaptivity} and thus enabling efficient LLM generation.

\section{ConfLayers}
\label{conflayers}

Building on what was previously discussed in Section \ref{related}, we introduce ConfLayers, a plug-and-play framework for dynamic confidence-based layer skipping in the context of self-speculative decoding. Figure \ref{fig6} demonstrates the generation process using ConfLayers. First, some input variables that control the algorithm such as the maximum number of optimization rounds and the acceptance rate goal for the search process are defined and the draft model is initialized as the result of uniformly skipping layers of the target model. 
Using that initial draft model, candidate tokens are generated and are then verified through the target model. After insuring that the conditions for the next step are met, the search for the best layer set to form the draft model then starts. The confidence of every layer is computed based on its Feed-Forward Neural Network (FFNN) and Attention module hidden states as explained in detail in Section \ref{layerskipping}. The confidence values at every layer are compared to the mean confidence over an adaptive window size that is decided per layer based on the aggressiveness of change in the confidence values and the index of the layer as elaborated in Section \ref{window}. The mean of the confidence values over the window is used as a threshold for the layer selection. If the confidence of a layer is less than the threshold, it is excluded from the draft model formation and added to the skip layer set \textit{\textbf{S}}. Once the set of layers to skip is found, the compact version of the model is used to form the draft model and generate the candidate tokens following the confidence-aware approach presented in the SWIFT framework \cite{xia2025swift}. The current set of non-skip layers is used to draft candidate tokens for a pre-defined search interval. Once that interval is reached, a search check is applied to evaluate the score of this search round. If the score of that round is greater the score goal or the maximum number of search rounds is reached, the search process is terminated and the best saved layer set \textit{\textbf{S}} is used as a draft model for the generation process of all remaining prompts. Otherwise, another round of search for the best draft model layer set is initiated.

\begin{figure}[t]
    \centering
    \includegraphics[width=\columnwidth]{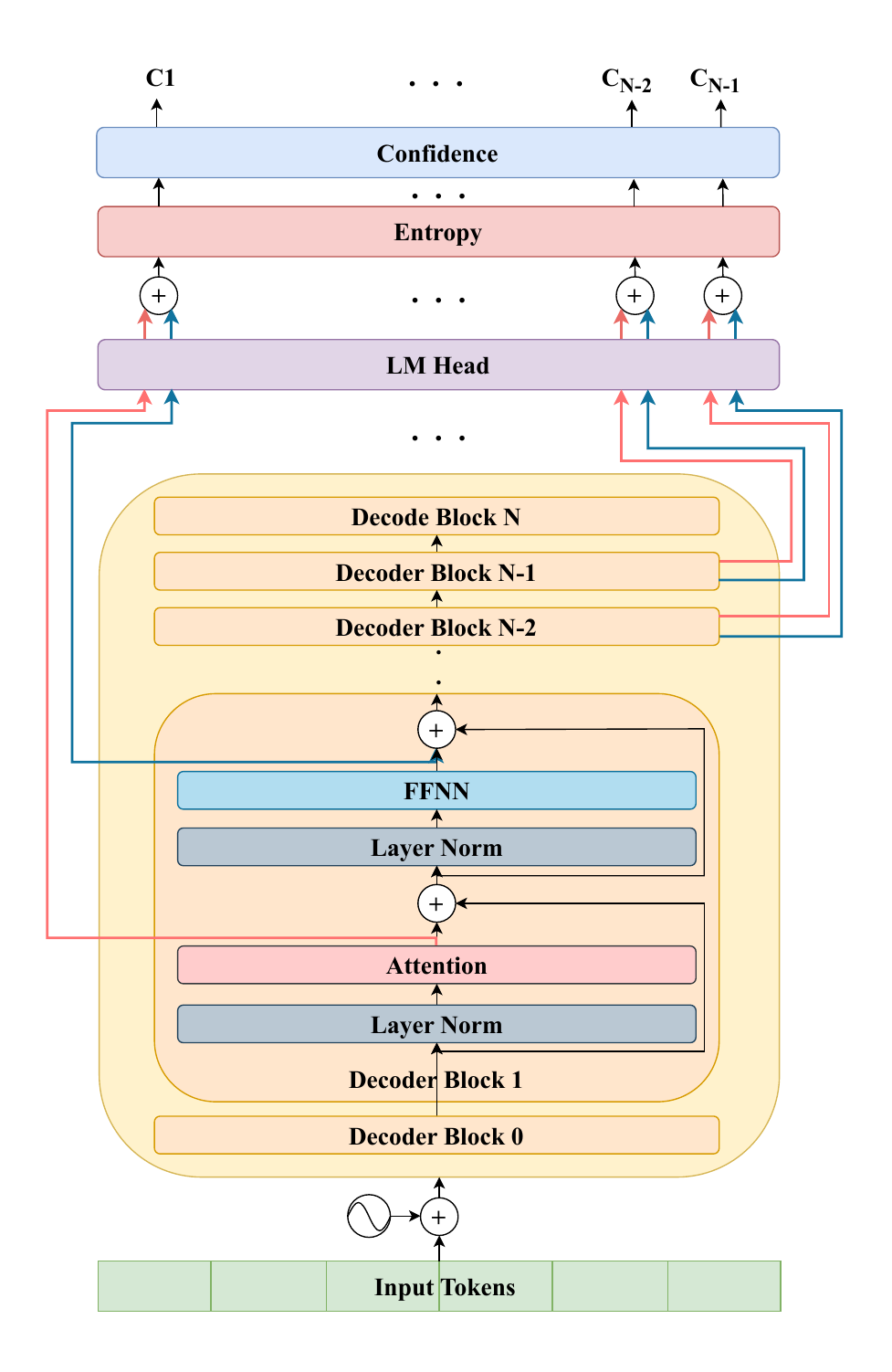}
    \caption{\centering Confidence computation detailed.}
    \label{fig8}
\end{figure}

\subsection{Confidence-based Layer Skipping}
\label{layerskipping}

The normalized confidence scores of the layers are found from the model logits by leveraging the relationship between predictive entropy and uncertainty. Figure \ref{fig8} demonstrates how the individual confidence values of every layer are found. It starts with extracting the hidden states of the FFNN and Attention modules.
The logits of every layer are obtained by projecting the FFNN and Attention hidden state outputs of that layer through the shared LM head. 
These projections are interpreted as layer-wise token distributions whose entropy reflects the uncertainty of the corresponding intermediate representation.
The raw logits are  casted to FP32 followed by a clamping operation to ensure stable numerical operations.
To derive the probability distribution of these stabilized logits, the softmax function is applied as seen in Equation \ref{eq:softmax}.

\begin{equation}
p_i = \frac{\exp(\ell_i)}{\sum_{j=1}^{K} \exp(\ell_j)}, \quad i = 1, \dots, K .
\label{eq:softmax}
\end{equation}

The uncertainty of this distribution is then quantified using Shannon entropy~\cite{shannon1948mathematical} in Equation \ref{eq:entropy}.

\begin{equation}
H(p) = - \sum_{i=1}^{K} p_i \log(p_i + \epsilon),
\label{eq:entropy}
\end{equation}

where $\epsilon$ is added for numerical stability when probabilities approach zero.
Finally, the confidence score is defined as the complement of normalized entropy in Equation \ref{eq:confidence}, where the maximum entropy corresponds to the uniform distribution and is given by Equation \ref{eq:maxentropy}.

\begin{equation}
\text{confidence} = 1 - \frac{H(p)}{H_{\max}} .
\label{eq:confidence}
\end{equation}

\begin{equation}
H_{\max} = \log K .
\label{eq:maxentropy}
\end{equation}

By construction, Equation \ref{eq:confidence} maps the entropy of the predictive distribution to the interval $[0,1]$, where values near $1$ indicate sharply peaked, confident predictions, while values near $0$ correspond to uncertain, near-uniform distributions.

\subsection{Adaptive Confidence-based Window Sizing for Layer Filtering}
\label{window}

With the confidence values of every layer now available, the proposed method identifies transformer layers suitable for skipping during inference based on confidence statistics. The decision-making process of including or excluding a layer in the draft model is based on comparing a layer's confidence value to the mean of its neighboring layers. The number of neighboring layers to include in this comparison is a window with an adaptive size based on the fluctuation in confidence values. 
Adaptive windowing improves the robustness of layer-skipping decisions by balancing sensitivity and stability in confidence comparisons. Fixed-size neighborhoods are either too noisy in regions with sharp confidence changes or too coarse in stable regions as we will see in Section \ref{ablation}, so adapting the window based on local confidence gradients and layer depth resolves this trade-off. 
This design aligns with the hierarchical nature of transformers, where early layers require fine-grained local comparison and later layers benefit from broader contextual aggregation. This also aligns with the need to adapt the comparison window to local confidence dynamics, using broader neighborhoods in regions with sharp confidence fluctuations for a more coarse-grained comparison and tighter neighborhoods where confidence evolves more smoothly  for a more fine-grained evaluation.

The input is a mapping from layer indices to confidence scores:
\[
\{ (l_1, c_1), (l_2, c_2), \ldots, (l_n, c_n) \},
\]
where $l_i$ denotes the $i$-th layer index and $c_i$ its associated confidence value.

To eliminate scale differences across layers, confidence values are standardized to zero mean and unit variance as shown in Equation \ref{eq:norm} where $\mu_c$ and $\sigma_c$ are the mean and standard deviation, and $\epsilon$ ensures numerical stability.

\begin{equation}
\hat{c}_i = \frac{c_i - \mu_c}{\sigma_c + \epsilon}
\label{eq:norm}
\end{equation}

To quantify the local fluctuations in confidence, we compute the second-order discrete gradient of the normalized sequence using Equation \ref{eq:grad} and rescale it into the range $[0, 1]$ via min--max normalization.

\begin{equation}
g = \nabla^2 \hat{c}
\label{eq:grad}
\end{equation} 

For each layer $i$, Equation \ref{eq:window} defines a local neighborhood window with a size adapting to the gradient magnitude and layer index, where $w_{\text{base}}$ is the static maximum number of neighboring layers to include, $l_i$ is the current layer index, and $num\_layers$ is the total number of layers in the model. $w_{\text{max}}$ is a variable defining the maximum additional number of layers to include which is reached if the gradient is zero at the last layer, signifying no change in the confidence for that layer. $w_{\text{max}}$ is proportional to the number of early consecutive layers that share the same confidence value before any change occurs which makes it adaptive to the model's confidence over the dataset in question.

\begin{equation}
w_i = w_{\text{base}} + w_{\text{max}} * g_i * (\frac{l_i}{num\_layers})
\label{eq:window}
\end{equation}
 
The window size varies with both the gradient and the index of the current layer. Stable regions (low $g_i$) yield smaller windows, while volatile regions (high $g_i$) yield larger ones. Simultaneously, earlier layers generate smaller windows and later layers generate larger windows. That is the case since the layer hidden states encode progressively higher-level abstractions, meaning that hidden states of early layers hold token-level lexical and syntactic information so only correlate with their immediate neighbors and a small window suffices. Later layers encode aggregate and stabilize contextual information and thus have long-range correlation in confidence, needing a larger window.

Excluding the current layer, the local mean and standard deviation are then computed as seen in Equations \ref{eq:mean} and \ref{eq:var}.

\begin{equation}
\mu_{i,\text{local}} = \text{mean}(\{\hat{c}_j : j \in w(i)\})
\label{eq:mean}
\end{equation}
\begin{equation}
\sigma_{i,\text{local}} = \text{std}(\{\hat{c}_j : j \in w(i)\})
\label{eq:var}
\end{equation}

A layer is marked for skipping if its confidence falls below the local baseline. We represent this condition in \eqref{eq:skip} with $\lambda$ controlling the threshold sensitivity. $\lambda$ is chosen on a per-model basis in a way to keep the number of layers to skip is between 40\% and 60\% as this is the range of skip layer set sizes that yield best results as explained in Appendix \ref{app4}.

\begin{equation}
\hat{c}_i < \mu_{i,\text{local}} - \lambda \cdot \sigma_{i,\text{local}}
\label{eq:skip}
\end{equation}

The algorithm outputs the set of all layers that satisfy this condition, sorted by index. By combining global normalization with local statistics, it adaptively identifies layers that contribute minimally to model confidence, thereby reducing computation during inference. We visualize an example of the relationship between the different variables described in this section in Appendix \ref{app2}.

\subsection{Interval-based Periodic Search}

Instead of initiating the search process after every generation cycle, a periodic mechanism shown in Figure \ref{fig6} is employed. Every non-skip layer set is used as a draft model for Opt\_Interval iterations where the number of accepted candidate tokens is accumulated at every iteration. After the end of every iteration, the search check ensures that the search process is halted once the maximum total number of iterations is reached. If that is not the case, the framework will check if the number of iterations in an interval have passed (Opt\_Interval). If the the number of iterations is still less that the number of iterations in an interval, the last best non-skip layer set saved is still adopted for the next generation cycle.
When the number of elapsed iterations reaches Opt\_Interval, the search process is triggered, and the iteration counter (Opt\_Iterations) is reset to begin a new period.
A final check is needed to examine if the average of the accumulated number of accepted tokens across the interval reaches a pre-defined minimum score signifying that the current set is good enough to adopt as the final non-skip layer set to form the draft model and no further search is needed, thus also halting the search process.

\begin{table*}[]
\caption{\centering Speedup provided by ConfLayers (CL) compared to DEL and SWIFT (SW) over vanilla LLM generation with several LLaMa models on different tasks}
\label{tab1}
\begin{tabular}{l|lll|lll|lll|lll}
\hline
Model           & \multicolumn{3}{l|}{\textbf{LLaMa-2-13B}}           & \multicolumn{3}{l|}{\textbf{LLaMa-2-70B}}           & \multicolumn{3}{l|}{\textbf{LLaMa-3-8B}}            & \multicolumn{3}{l}{\textbf{LLaMa-3-70B}}            \\ \hline
Approach        & \textbf{DEL} & \textbf{SW}  & \textbf{CL}           & \textbf{DEL} & \textbf{SW}  & \textbf{CL}           & \textbf{DEL} & \textbf{SW}  & \textbf{CL}           & \textbf{DEL} & \textbf{SW}  & \textbf{CL}           \\ \hline
\textbf{CNN-DM} & 0.89$\times$ & 0.92$\times$ & \textbf{1.16$\times$} & 0.95$\times$ & 1.30$\times$ & \textbf{1.37$\times$} & 0.77$\times$ & 1.08$\times$ & \textbf{1.10$\times$} & 0.89$\times$ & 1.26$\times$ & \textbf{1.38$\times$} \\
\textbf{GSM8K}  & 0.89$\times$ & 1.09$\times$ & \textbf{1.13$\times$} & 0.92$\times$ & 1.31$\times$ & \textbf{1.34$\times$} & 0.74$\times$ & 1.09$\times$ & \textbf{1.10$\times$} & 0.85$\times$ & 1.23$\times$ & \textbf{1.33$\times$} \\
\textbf{WMT14}  & 0.99$\times$ & 1.07$\times$ & \textbf{1.17$\times$} & 0.90$\times$ & 1.11$\times$ & \textbf{1.35$\times$} & 0.73$\times$ & 1.02$\times$ & \textbf{1.05$\times$} & 0.89$\times$ & 1.28$\times$ & \textbf{1.35$\times$} \\
\textbf{Alpaca} & 0.95$\times$ & 1.04$\times$ & \textbf{1.13$\times$} & 0.89$\times$ & 1.23$\times$ & \textbf{1.35$\times$} & 0.74$\times$ & 1.05$\times$ & \textbf{1.19$\times$} & 0.84$\times$ & 1.06$\times$ & \textbf{1.16$\times$} \\ \hline
Average & 0.93$\times$ & 1.03$\times$ & \textbf{1.15$\times$} & 0.92$\times$ & 1.24$\times$ & \textbf{1.35$\times$} & 0.75$\times$ & 1.06$\times$ & \textbf{1.11$\times$} & 0.87$\times$ & 1.21$\times$ & \textbf{1.31$\times$} \\ \hline
\end{tabular}
\end{table*}

\section{Evaluation}
\label{eval}

\subsection{Experimental Setup}
\label{expsetup}

\subsubsection{Inference Setup}
We evaluate the performance of ConfLayers across a variety of model types, model sizes and tasks.
For the main evaluation, the models include LLaMa-2 \cite{touvron2023llama2openfoundation} of sizes 13B and 70B and LLaMa-3 \cite{llama3modelcard} of sizes 8B and 70B. The models are loaded with FP16 precision. The datasets used for this evaluation are CNN/Daily Mail (CNN-DM) for summarization, GSM8K for mathematical reasoning, WMT14/DE-EN for translation, and Alpaca for instruction.
We conduct 0-shot evaluation for translation and instruction, 1-shot evaluation for summarization and 5-shot evaluation for mathematical reasoning.
We also evaluate the performance over task-specific models such as CodeLLaMa \cite{rozière2024codellamaopenfoundation} for code generation with 34B parameters using the HumanEval dataset and Qwen-2.5-Math \cite{yang2024qwen25mathtechnicalreportmathematical} with 72B parameters for mathematical reasoning using the GSM8K dataset.

The base window $ w_{\text{base}}$ is set to 2 to provide a minimal yet stable context for confidence estimation, and the value of the threshold sensitivity $\lambda$ is chosen per-model with $0.25<\lambda<0.5$ (as explained in Section \ref{analysis}). We set the threshold score to halt the search to four accepted tokens following common practice in state-of-the-art self-speculative decoding frameworks, the minimum skip ratio to 40\%, the context window to 100, and the search interval to 30. These values were selected through sensitivity analysis to balance reliability, adaptivity, and overhead in the search process.

We compare our approach to DEL ~\cite{zarch2025del} and SWIFT \cite{xia2025swift}, two other plug-and-play approaches for self-speculative decoding. To setup the frameworks, we adopt the descriptions provided in their respective papers.
The baseline for this evaluation is vanilla auto-regressive decoding without acceleration.
We report the speedup \textbf{$s$} of all frameworks over vanilla LLM generation. 
The generation is evaluated over 100 samples with the maximum number of tokens generated set to 512 tokens on an AMD Instinct MI300X GPU running ROCm 6.4.1.

\subsubsection{Metrics}

We report the speedup observed with ConfLayers (referred to as CL in the table), DEL and SWIFT compared to vanilla LLM generation, the acceptance rate $\alpha$ defined as the ratio of the number of tokens accepted by the target model over the total number of drafted tokens, the skip rate $\beta$ representing the percentage of layers skipped from the total number of layers to form the draft model, and the number of accepted drafted tokens $M$ following the target model validation. Results for the last two metrics are reported as the mean value across all test samples. To quantify accuracy and output quality, we report the Rouge-2 metric \cite{ganesan2015rouge} evaluating similarity between the output of the self-speculative approach and the vanilla generation one.

\subsection{Results}
\label{results}

\subsubsection{Speedup Results for LLaMa Models}
Table \ref{tab1} presents the main results of our evaluation on the four LLaMa models. 
The results show that ConfLayers always achieves a speedup above 1 and up to 1.4$\times$, proving that our approach can accelerate LLM generation and provide an advantage over the vanilla approach. The table also shows that ConfLayers consistently achieves significantly higher speedup results than DEL and SWIFT, especially for larger models such as LLaMa-2-70B and LLaMa-3-70B. DEL fails to provide any speedup over the vanilla autoregressive generation as it requires the models to be fine-tuned for LayerSkip-based early exiting. ConfLayers on the other hand does not require any fine-tuning and is able to accelerate the generation for any model. When the larger LLaMa models are used for a summarization task over the CNN-DM dataset, our approach can yield a 1.37$\times$ and 1.38$\times$ speedup respectively compared to vanilla LLM generation while SWIFT can only reach a 1.30$\times$ and 1.26$\times$ speedup. Additionally, while SWIFT struggles to achieve speedup in cases like inference with LLaMa-2-13B on CNN-DM, ConfLayers is still able to accelerate inference by 1.16$\times$. We present more a detailed comparison between ConfLayers' and SWIFT's speedup results in Appendix \ref{app1}. We also present a visualization of the progression of the skipped layer set throughout the search process in Appendix \ref{app3}.

\begin{figure*}[]
    \centering
    \includegraphics[width=\textwidth]{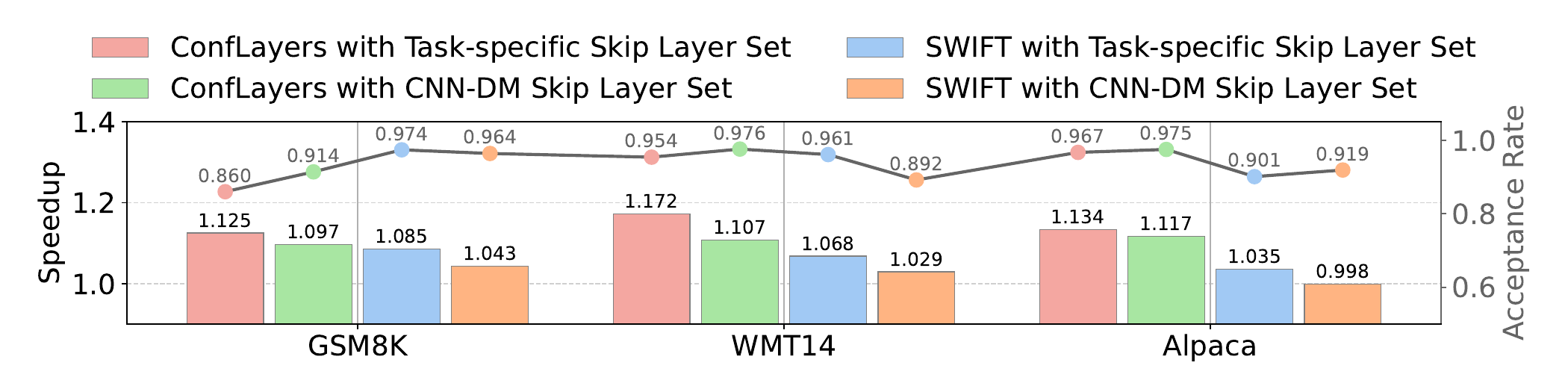}
    \caption{\centering Speedup comparison between SWIFT and ConfLayers with LLaMa-2-13B on a dynamic input stream.}
    \label{fig14}
\end{figure*}

\subsubsection{Results for Specialized Models}

We expand the evaluation to include specialized models of different sizes such as CodeLLaMa-34B and Qwen2.5-Math-72B and present the results in Table \ref{tab2}. ConfLayers outperforms the vanilla LLM generation by 1.24$\times$ and 1.22$\times$ for inference with CodeLLaMa-34B on the HumanEval coding dataset and Qwen2.5-Math-72B on the GSM8K mathematical reasoning dataset, respectively. Our approach also outperforms SWIFT which provides a 1.06$\times$ speedup for inference with CodeLLaMa-34B and a speedup of 1.15$\times$ for inference with Qwen2.5-Math-72B with a remarkably lower number of accepted tokens than ConfLayers in both cases. Finally, the $Rouge-2$ results fall in high overlap ranges relative to the tasks deployed, which expected from speculative decoding methods adopted with greedy sampling as they conserve the probability distribution of the target model. This observation supports the claim that ConfLayers preserves the output quality of the target model.

\subsubsection{Performance Discussion}
\begin{figure}[t]
    \centering
    \includegraphics[width=\columnwidth]{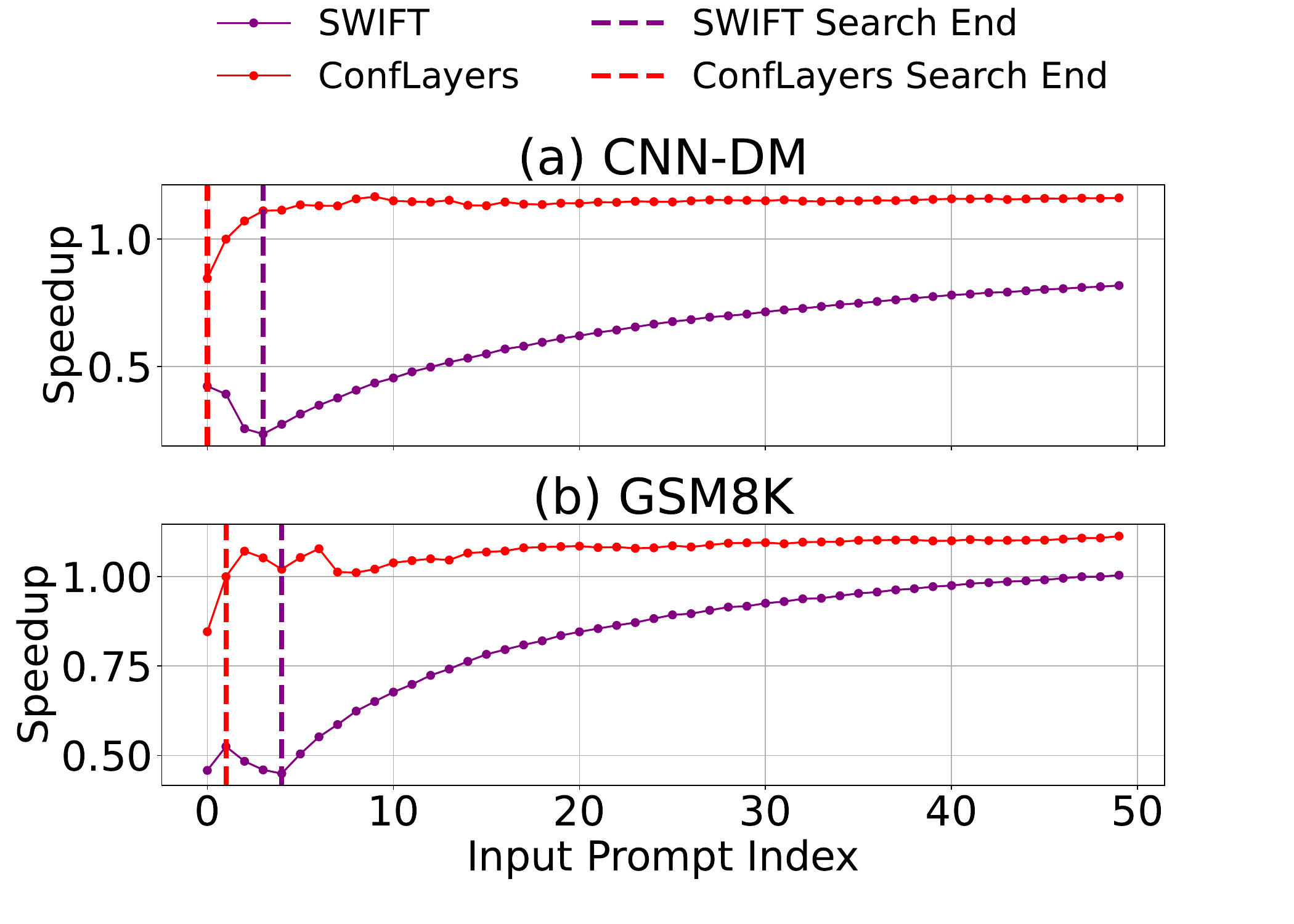}
    \caption{\centering Speedup at every input prompt for ConfLayers and SWIFT with LLaMa-2-13B on (a) CNN-DM and (b) GSM8K showing that ConfLayers achieves high speedup even in the early stages of the search process, unlike SWIFT.}
    \label{fig9}
\end{figure}

The observed improvement is due to a key difference between ConfLayers and SWIFT, which is the objective function. ConfLayers prioritizes maximizing the number of accepted tokens by the target model instead of prioritizing optimizing for the acceptance rate as SWIFT does, since a draft model with high acceptance rate and low number of accepted tokens might not be the best performing. We notice that manifesting in results with CodeLLaMa, as the acceptance rate when ConfLayers is adopted is slightly lower than the one measured with SWIFT, but ConfLayers is still able to provide a higher speedup as it can achieve a higher number of accepted tokens compared to SWIFT. An additional helping factor is the flexibility that ConfLayers provides during the search process. While SWIFT only allows a fixed skipping rate, Conflayers does not limit the search space to a certain number of layers to skip and expands the exploration set to any number of layers below or above a pre-defined threshold while conditioning the search on the performance of the draft model. Additionally, ConfLayers provides significant speedup early on, even during the search process, as the initial skip layer sets  unlike SWIFT's initial sets as shown in Figure \ref{fig9}.
Thus, although ConfLayers' acceptance rates can be marginally lower than SWIFT’s, our approach is still able to outperform SWIFT due to the ability of ConfLayers to quickly find a smaller draft model that preserves the quality of the generated tokens. Figure \ref{fig9} also shows that the search process does not add any significant overhead to the inference, as the speedup observed for both tasks is equal or above 1. That is due to ConfLayers' design where the search is performed online during inference. As a result, confidence estimation does not introduce a separate search phase, and its overhead is limited to lightweight operations and is amortized by the layers skipped and does not measurably affect end-to-end latency.

\begin{table}[]
\centering
\caption{Performance evaluation of ConfLayers compared to SWIFT and vanilla LLM generation with specialized models}
\label{tab2}
\begin{tabular}{l|ll|ll}
\hline
\textbf{Model}     & \multicolumn{2}{l|}{\textbf{CodeLLaMa-34B}} & \multicolumn{2}{l}{\textbf{Qwen2.5-Math-72B}} \\ \hline
\textbf{Dataset}   & \multicolumn{2}{l|}{\textbf{HumanEval}}     & \multicolumn{2}{l}{\textbf{GSM8K}}            \\ \hline
\textbf{Approach}  & \textbf{SWIFT}    & \textbf{CL}             & \textbf{SWIFT}     & \textbf{CL}              \\ \hline
\textbf{$Speedup$}       & 1.06$\times$      & \textbf{1.24$\times$}   & 1.15$\times$       & \textbf{1.22$\times$}    \\
\textbf{$\alpha$}  & 0.97              & 0.94                    & 0.78               & 0.84                     \\
\textbf{$\beta$}   & 50\%              & 44\%                    & 50\%               & 44\%                     \\
\textbf{$M$}       & 3.76              & 4.65                    & 2.64               & 3.47                     \\
\textbf{$Rouge-2$} & 0.057             & 0.054                   & 0.029              & 0.032                    \\ \hline
\end{tabular}
\end{table}

\subsection{In-Depth Analysis}
\label{analysis}

\subsubsection{Dynamic Input Stream}

We evaluate ConfLayers' ability to adapt to a dynamic input stream where the same skip layer set is adopted to run inference on several different tasks. Instead of searching for the best layer set for every dataset, we only search for the set on one of the datasets and deploy that set when running inference on the rest of them to examine the framework's ability to adapt to other tasks. In this study, we find the best layer set with LLaMa-2-13B on the summarization task (CNN-DM) and deploy it on 100 samples of each of the mathematical reasoning dataset (GSM8K), the english-to-german translation dataset (WMT14), and the instruction dataset (Alpaca). We also implement the same approach with SWIFT.
We evaluate the performance of the framework and report the speedup and acceptance rate findings in Figure \ref{fig14}. We compare these results to the ones obtained by optimizing the skip layer set for each task separately. Across all datasets, we notice that ConfLayers conserves its high performance when the layer set of another task is adopted during inference as it consistently provides a speedup greater or equal than 1.1$\times$ on all three tasks. We also observe that the version of ConfLayers with the dynamic input stream does not only perform better than the SWIFT framework with dynamic input but even outperforms SWIFT in the regular case where the skip layer set is obtained for every task separately. This could be easily seen when running inference on the Alpaca dataset as Conflayers achieves a speedup of 1.134$\times$ and 1.117$\times$ with task-specific and dynamic skip layer sets respectively, while SWIFT can only achieve a 1.035$\times$ with its task-specific skip layer set and fails to achieve speedup with the dynamic input resulting in a speedup of 0.998$\times$.
ConfLayers' conservation of performance also manifests when a dynamic input stream is adopted by consistently providing an equal or greater token acceptance rate compared to ConfLayers with a regular task-separated input. A clear example of that is the increase in acceptance rate from 0.954 to 0.976 observed when the CNN-DM skip layer set is deployed for inference on the WMT14 dataset with ConfLayers, while SWIFT shows a drop in acceptance rate from 0.961 to 0.892 for the same setup.

\begin{figure}[t]
    \centering
    \includegraphics[width=\columnwidth]{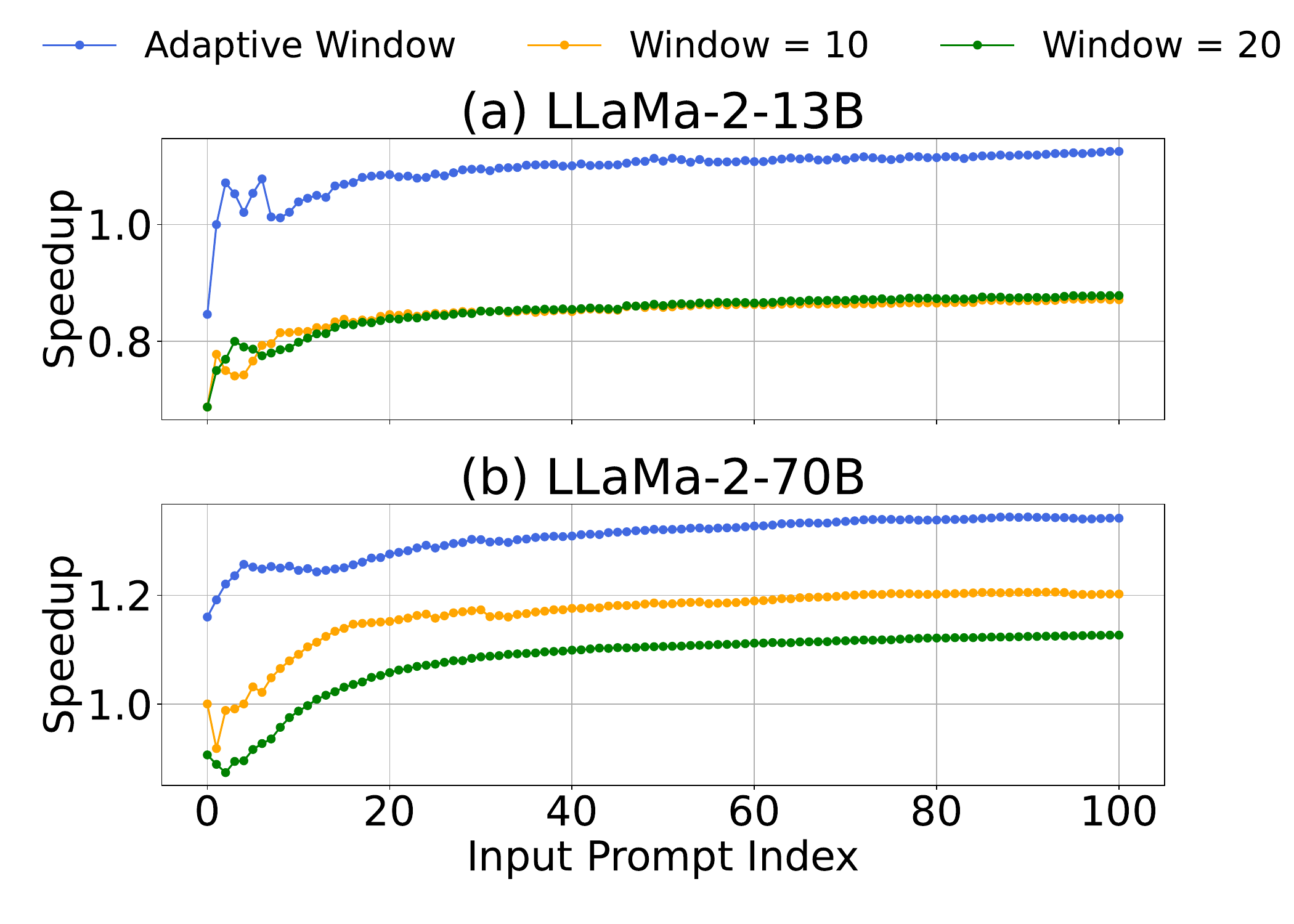}
    \caption{\centering Speedup at every input prompt for different windowing techniques on (a) LLaMa-2-13B and (b) LLaMa-2-70B.}
    \label{fig4}
\end{figure}

\subsubsection{Ablation Study: Adaptive Filtering}
\label{ablation}

To highlight the importance of dynamic window sizing for layer filtering previously discussed in Section \ref{window}, we conduct an ablation study of this feature using the LLaMa-2-13B and LLaMa-2-70B models on GSM8K and report the results in Figures \ref{fig4} (a) and \ref{fig4} (b) respectively. We set the window size to constant values of 10 and 20 for all layers and compare the resulting performance to the one we get when the adaptive window technique is adopted. We notice that, in Figure \ref{fig4} (a), the generation with LLaMa-2-13B performs poorly in both cases where the window size is set to 10 and 20 across all layers as the resulting speedup is less than 0.9$\times$. This shows that using a constant window size across all layers fails to deliver an improvement over vanilla generation. Although we observe better results when a constant window size is adopted When inference is run with LLaMa-2-70B, we notice an inconsistency across models. Figure \ref{fig4} (b) shows that drafting tokens with a constant window of ten layers offers a ~1.2$\times$ speedup compared to a ~1.5$\times$ speedup when skipping layers with a constant window size of 20 layers. However, this shows that the performance of the framework with different constant window sizes is not consistent across LLaMa-2-13B and LLaMa-2-70B. The poor and inconsistent performance proves the need for an adaptive technique to pick the window size.
Figures \ref{fig4} (a) and (b) clearly show that the generation with the adaptive window technique performs significantly better than generation with both constant window sizes, offering a speedup greater than 1.1$\times$ and 1.3$\times$ with LLaMa-2-13B and LLaMa-2-70B respectively.

\section{Conclusion}
\label{conclusion}

We introduced ConfLayers, a confidence-based method for constructing draft models in self-speculative decoding. By adaptively skipping layers based on intermediate confidence scores, ConfLayers eliminates the need for training a policy while preserving strong speed–quality trade-offs. Across multiple models and datasets, our approach achieves up to 1.4$\times$ faster inference compared to vanilla generation. These results demonstrate that simple, confidence-guided heuristics can be an effective alternative to learning-based methods, offering a practical step toward more efficient large language model inference.

\bibliography{Conflayers}

@inproceedings{leviathan2023specdec,
author = {Leviathan, Yaniv and Kalman, Matan and Matias, Yossi},
title = {Fast inference from transformers via speculative decoding},
year = {2023},
publisher = {JMLR.org},
booktitle = {Proceedings of the 40th International Conference on Machine Learning},
articleno = {795},
numpages = {13},
location = {Honolulu, Hawaii, USA},
series = {ICML'23}
}

@misc{chen2023specdec,
      title={Accelerating Large Language Model Decoding with Speculative Sampling}, 
      author={Charlie Chen and Sebastian Borgeaud and Geoffrey Irving and Jean-Baptiste Lespiau and Laurent Sifre and John Jumper},
      year={2023},
      eprint={2302.01318},
      archivePrefix={arXiv},
      primaryClass={cs.CL},
      url={https://arxiv.org/abs/2302.01318}, 
}

@inproceedings{li2024eagle,
author = {Li, Yuhui and Wei, Fangyun and Zhang, Chao and Zhang, Hongyang},
title = {EAGLE: speculative sampling requires rethinking feature uncertainty},
year = {2024},
publisher = {JMLR.org},
booktitle = {Proceedings of the 41st International Conference on Machine Learning},
articleno = {1162},
numpages = {14},
location = {Vienna, Austria},
series = {ICML'24}
}

@inproceedings{zhang2024draftverify,
    title = "Draft {\&} Verify: Lossless Large Language Model Acceleration via Self-Speculative Decoding",
    author = "Zhang, Jun  and
      Wang, Jue  and
      Li, Huan  and
      Shou, Lidan  and
      Chen, Ke  and
      Chen, Gang  and
      Mehrotra, Sharad",
    editor = "Ku, Lun-Wei  and
      Martins, Andre  and
      Srikumar, Vivek",
    booktitle = "Proceedings of the 62nd Annual Meeting of the Association for Computational Linguistics (Volume 1: Long Papers)",
    month = aug,
    year = "2024",
    address = "Bangkok, Thailand",
    publisher = "Association for Computational Linguistics",
    url = "https://aclanthology.org/2024.acl-long.607/",
    doi = "10.18653/v1/2024.acl-long.607",
    pages = "11263--11282"
}

@inproceedings{metel2024draftonthefly,
    title = "Draft on the Fly: Adaptive Self-Speculative Decoding using Cosine Similarity",
    author = "Metel, Michael R.  and
      Lu, Peng  and
      Chen, Boxing  and
      Rezagholizadeh, Mehdi  and
      Kobyzev, Ivan",
    editor = "Al-Onaizan, Yaser  and
      Bansal, Mohit  and
      Chen, Yun-Nung",
    booktitle = "Findings of the Association for Computational Linguistics: EMNLP 2024",
    month = nov,
    year = "2024",
    address = "Miami, Florida, USA",
    publisher = "Association for Computational Linguistics",
    url = "https://aclanthology.org/2024.findings-emnlp.124/",
    doi = "10.18653/v1/2024.findings-emnlp.124",
    pages = "2267--2272"
}

@inproceedings{elhoushi2024layerskip,
    title = "{L}ayer{S}kip: Enabling Early Exit Inference and Self-Speculative Decoding",
    author = "Elhoushi, Mostafa  and
      Shrivastava, Akshat  and
      Liskovich, Diana  and
      Hosmer, Basil  and
      Wasti, Bram  and
      Lai, Liangzhen  and
      Mahmoud, Anas  and
      Acun, Bilge  and
      Agarwal, Saurabh  and
      Roman, Ahmed  and
      Aly, Ahmed  and
      Chen, Beidi  and
      Wu, Carole-Jean",
    editor = "Ku, Lun-Wei  and
      Martins, Andre  and
      Srikumar, Vivek",
    booktitle = "Proceedings of the 62nd Annual Meeting of the Association for Computational Linguistics (Volume 1: Long Papers)",
    month = aug,
    year = "2024",
    address = "Bangkok, Thailand",
    publisher = "Association for Computational Linguistics",
    url = "https://aclanthology.org/2024.acl-long.681/",
    doi = "10.18653/v1/2024.acl-long.681",
    pages = "12622--12642"
}

@inproceedings{liu2024kangaroo,
 author = {Liu, Fangcheng and Tang, Yehui and Liu, Zhenhua and Ni, Yunsheng and Tang, Duyu and Han, Kai and Wang, Yunhe},
 booktitle = {Advances in Neural Information Processing Systems},
 editor = {A. Globerson and L. Mackey and D. Belgrave and A. Fan and U. Paquet and J. Tomczak and C. Zhang},
 pages = {11946--11965},
 publisher = {Curran Associates, Inc.},
 title = {Kangaroo: Lossless Self-Speculative Decoding for Accelerating LLMs via Double Early Exiting},
 url = {https://proceedings.neurips.cc/paper_files/paper/2024/file/16336d94a5ffca8de019087ab7fe403f-Paper\\-Conference.pdf},
 volume = {37},
 year = {2024}
}

@inproceedings{
xia2025swift,
title={{SWIFT}: On-the-Fly Self-Speculative Decoding for {LLM} Inference Acceleration},
author={Heming Xia and Yongqi Li and Jun Zhang and Cunxiao Du and Wenjie Li},
booktitle={The Thirteenth International Conference on Learning Representations},
year={2025},
url={https://openreview.net/forum?id=EKJhH5D5wA}
}

@inproceedings{chen2025clasp,
    title = "{CL}a{S}p: In-Context Layer Skip for Self-Speculative Decoding",
    author = "Chen, Longze  and
      Shan, Renke  and
      Wang, Huiming  and
      Wang, Lu  and
      Liu, Ziqiang  and
      Luo, Run  and
      Wang, Jiawei  and
      Alinejad-Rokny, Hamid  and
      Yang, Min",
    editor = "Che, Wanxiang  and
      Nabende, Joyce  and
      Shutova, Ekaterina  and
      Pilehvar, Mohammad Taher",
    booktitle = "Proceedings of the 63rd Annual Meeting of the Association for Computational Linguistics (Volume 1: Long Papers)",
    month = jul,
    year = "2025",
    address = "Vienna, Austria",
    publisher = "Association for Computational Linguistics",
    url = "https://aclanthology.org/2025.acl-long.1525/",
    doi = "10.18653/v1/2025.acl-long.1525",
    pages = "31608--31618",
    ISBN = "979-8-89176-251-0"
}

@inproceedings{
zarch2025del,
title={{DEL}: Context-Aware Dynamic Exit Layer for Efficient Self-Speculative Decoding},
author={Hossein Entezari Zarch and Lei Gao and Chaoyi Jiang and Murali Annavaram},
booktitle={Second Conference on Language Modeling},
year={2025},
url={https://openreview.net/forum?id=cAFxSuXQvT}
}

@misc{song2025knnssd,
      title={KNN-SSD: Enabling Dynamic Self-Speculative Decoding via Nearest Neighbor Layer Set Optimization}, 
      author={Mingbo Song and Heming Xia and Jun Zhang and Chak Tou Leong and Qiancheng Xu and Wenjie Li and Sujian Li},
      year={2025},
      eprint={2505.16162},
      archivePrefix={arXiv},
      primaryClass={cs.CL},
      url={https://arxiv.org/abs/2505.16162}, 
}

@misc{rozière2024codellamaopenfoundation,
      title={Code Llama: Open Foundation Models for Code}, 
      author={Baptiste Rozière and Jonas Gehring and Fabian Gloeckle and Sten Sootla and Itai Gat and Xiaoqing Ellen Tan and Yossi Adi and Jingyu Liu and Romain Sauvestre and Tal Remez and Jérémy Rapin and Artyom Kozhevnikov and Ivan Evtimov and Joanna Bitton and Manish Bhatt and Cristian Canton Ferrer and Aaron Grattafiori and Wenhan Xiong and Alexandre Défossez and Jade Copet and Faisal Azhar and Hugo Touvron and Louis Martin and Nicolas Usunier and Thomas Scialom and Gabriel Synnaeve},
      year={2024},
      eprint={2308.12950},
      archivePrefix={arXiv},
      primaryClass={cs.CL},
      url={https://arxiv.org/abs/2308.12950}, 
}

@article{yang2024qwen25mathtechnicalreportmathematical,
  title={Qwen2.5-Math Technical Report: Toward Mathematical Expert Model via Self-Improvement}, 
  author={An Yang and Beichen Zhang and Binyuan Hui and Bofei Gao and Bowen Yu and Chengpeng Li and Dayiheng Liu and Jianhong Tu and Jingren Zhou and Junyang Lin and Keming Lu and Mingfeng Xue and Runji Lin and Tianyu Liu and Xingzhang Ren and Zhenru Zhang},
  journal={arXiv preprint arXiv:2409.12122},
  year={2024}
}

@misc{touvron2023llama2openfoundation,
      title={Llama 2: Open Foundation and Fine-Tuned Chat Models}, 
      author={Hugo Touvron and Louis Martin and Kevin Stone and Peter Albert and Amjad Almahairi and Yasmine Babaei and Nikolay Bashlykov and Soumya Batra and Prajjwal Bhargava and Shruti Bhosale and Dan Bikel and Lukas Blecher and Cristian Canton Ferrer and Moya Chen and Guillem Cucurull and David Esiobu and Jude Fernandes and Jeremy Fu and Wenyin Fu and Brian Fuller and Cynthia Gao and Vedanuj Goswami and Naman Goyal and Anthony Hartshorn and Saghar Hosseini and Rui Hou and Hakan Inan and Marcin Kardas and Viktor Kerkez and Madian Khabsa and Isabel Kloumann and Artem Korenev and Punit Singh Koura and Marie-Anne Lachaux and Thibaut Lavril and Jenya Lee and Diana Liskovich and Yinghai Lu and Yuning Mao and Xavier Martinet and Todor Mihaylov and Pushkar Mishra and Igor Molybog and Yixin Nie and Andrew Poulton and Jeremy Reizenstein and Rashi Rungta and Kalyan Saladi and Alan Schelten and Ruan Silva and Eric Michael Smith and Ranjan Subramanian and Xiaoqing Ellen Tan and Binh Tang and Ross Taylor and Adina Williams and Jian Xiang Kuan and Puxin Xu and Zheng Yan and Iliyan Zarov and Yuchen Zhang and Angela Fan and Melanie Kambadur and Sharan Narang and Aurelien Rodriguez and Robert Stojnic and Sergey Edunov and Thomas Scialom},
      year={2023},
      eprint={2307.09288},
      archivePrefix={arXiv},
      primaryClass={cs.CL},
      url={https://arxiv.org/abs/2307.09288}, 
}

@article{llama3modelcard,
title={Llama 3 Model Card},
author={AI@Meta},
year={2024},
url = {https://github.com/meta-llama/llama3/blob/main/MODEL_CARD.md}
}

@article{shannon1948mathematical,
  title={A Mathematical Theory of Communication},
  author={Shannon, Claude E},
  journal={Bell System Technical Journal},
  volume={27},
  number={3},
  pages={379--423},
  year={1948}
}

@inproceedings{
chen2024sequoia,
title={Sequoia: Scalable and Robust Speculative Decoding},
author={Zhuoming Chen and Avner May and Ruslan Svirschevski and Yu-Hsun Huang and Max Ryabinin and Zhihao Jia and Beidi Chen},
booktitle={The Thirty-eighth Annual Conference on Neural Information Processing Systems},
year={2024},
url={https://openreview.net/forum?id=rk2L9YGDi2}
}

@inproceedings{
timor2025lossless,
title={Accelerating {LLM} Inference with Lossless Speculative Decoding Algorithms for Heterogeneous Vocabularies},
author={Nadav Timor and Jonathan Mamou and Daniel Korat and Moshe Berchansky and Gaurav Jain and Oren Pereg and Moshe Wasserblat and David Harel},
booktitle={Forty-second International Conference on Machine Learning},
year={2025},
url={https://openreview.net/forum?id=vQubr1uBUw}
}

@inproceedings{
liao2025rewardguided,
title={Reward-Guided Speculative Decoding for Efficient {LLM} Reasoning},
author={Baohao Liao and Yuhui Xu and Hanze Dong and Junnan Li and Christof Monz and Silvio Savarese and Doyen Sahoo and Caiming Xiong},
booktitle={Forty-second International Conference on Machine Learning},
year={2025},
url={https://openreview.net/forum?id=AVeskAAETB}
}

@inproceedings{
li2024nearest,
title={Nearest Neighbor Speculative Decoding for {LLM} Generation and Attribution},
author={Minghan Li and Xilun Chen and Ari Holtzman and Beidi Chen and Jimmy Lin and Wen-tau Yih and Xi Victoria Lin},
booktitle={The Thirty-eighth Annual Conference on Neural Information Processing Systems},
year={2024},
url={https://openreview.net/forum?id=Ni9kebsSTt}
}

@inproceedings{
anonymous2026speculative,
title={Speculative Speculative Decoding},
author={Anonymous},
booktitle={The Fourteenth International Conference on Learning Representations},
year={2026},
url={https://openreview.net/forum?id=aL1Wnml9Ef}
}

@article{ganesan2015rouge,
  title={ROUGE 2.0: Updated and Improved Measures for Evaluation of Summarization Tasks},
  author={Ganesan, Kavita},
  year={2015}
}

@inproceedings{li2025multisample,
    title = "Speculative Decoding for Multi-Sample Inference",
    author = "Li, Yiwei  and
      Shi, Jiayi  and
      Feng, Shaoxiong  and
      Yuan, Peiwen  and
      Wang, Xinglin  and
      Zhang, Yueqi  and
      Zhang, Ji  and
      Tan, Chuyi  and
      Pan, Boyuan  and
      Hu, Yao  and
      Li, Kan",
    editor = "Christodoulopoulos, Christos  and
      Chakraborty, Tanmoy  and
      Rose, Carolyn  and
      Peng, Violet",
    booktitle = "Findings of the Association for Computational Linguistics: EMNLP 2025",
    month = nov,
    year = "2025",
    address = "Suzhou, China",
    publisher = "Association for Computational Linguistics",
    url = "https://aclanthology.org/2025.findings-emnlp.668/",
    doi = "10.18653/v1/2025.findings-emnlp.668",
    pages = "12523--12533",
    ISBN = "979-8-89176-335-7"
}
\bibliographystyle{icml2026}

\newpage
\appendix
\onecolumn

\section{Impact of $\lambda$ and \textbf{$\beta$} on Layer Selection}
\label{app4}

\begin{figure}[h]
    \centering
    \includegraphics[width=.7\columnwidth]{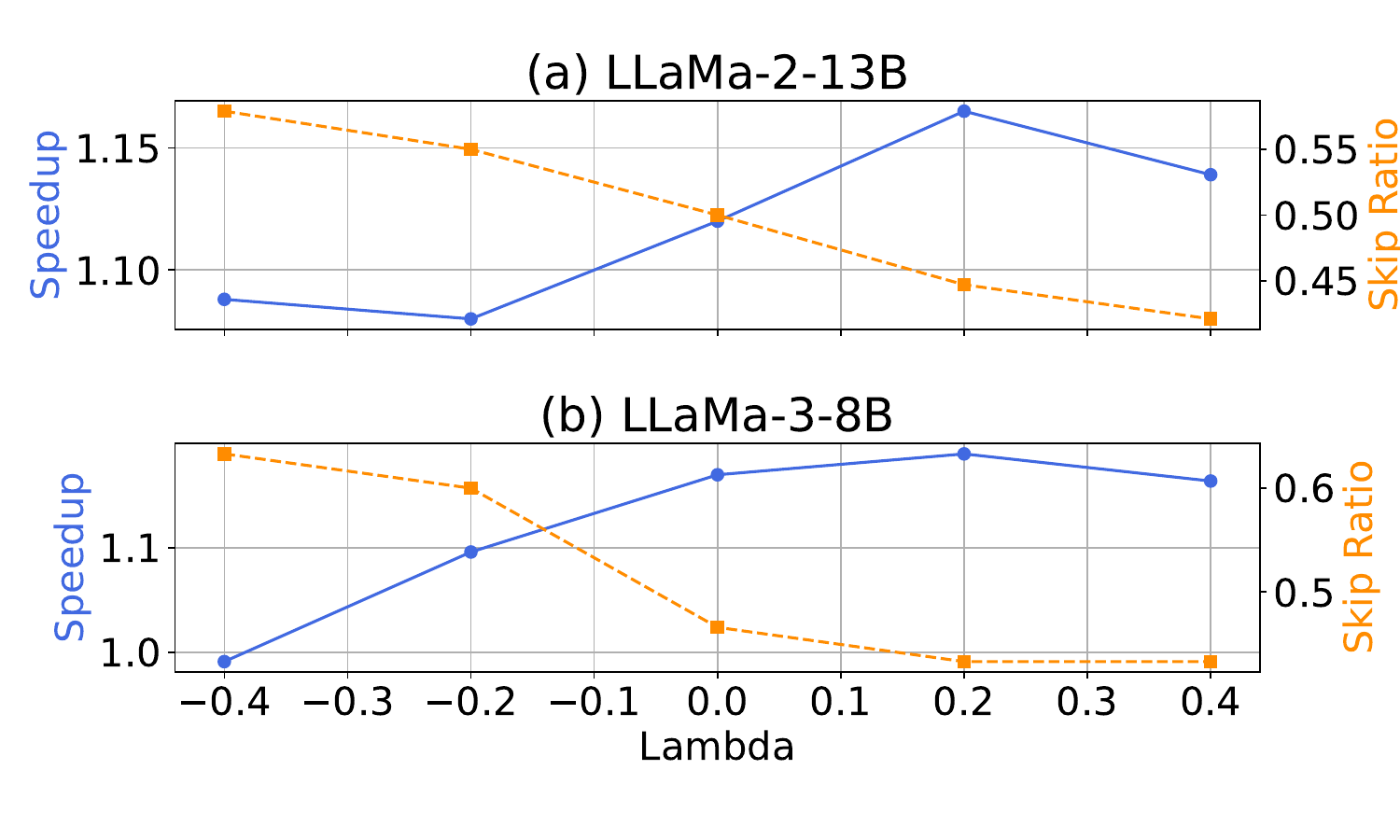}
    \caption{\centering Speedup at every input prompt for different $\lambda$ values on Alpaca with (a) LLaMa-2-13B and (b) LLaMa-3-8B.}
    \label{fig13}
\end{figure}

The number of layers in the skip layer set can impact the performance of the draft model. However, since ConfLayers does not restrict the search process to a certain skip ratio, the latter is controlled by varying the minimum permissible skip ratio alongside $\lambda$. This parameter will directly affect the range of layers that satisfy the condition in \eqref{eq:skip} and join the skip layer set, and thus directly impact the skip ratio range that is considered part of the search space. Thus, $\lambda$ is a key parameter that controls the sensitivity threshold and consequently the skip ratio. 
Intuitively, the framework should perform better as lambda increases and as the skip ratio decreases until its performance hits a peak then starts dropping again as the skip ratio further decreases, causing the draft model to increase in size and lose its advantage.
To study and quantify the impact of $\lambda$ on our approach, we examine the speedup and skip ratio when varying $\lambda$ with LLaMa-2-13B and LLaMa-3-8B on 50 prompt samples of the Alpaca dataset. The minimum permissible skip ratio is set to 40\% in all of the cases. 
The results visualized in Figure \ref{fig13} 
support the claim previously made about the relationship between $\lambda$ and the performance as we can see that the speedup increases when $\lambda$ increases and the skip ratio decreases for both models. The performance peaks at a skip ratio between 40\% and 50\% in both cases, so we adopt a skip ratio in that range when running the main result evaluation for all models. 

\section{Correlation Visualization Example}
\label{app2}

\begin{figure}[h]
    \centering
    \includegraphics[width=.6\columnwidth]{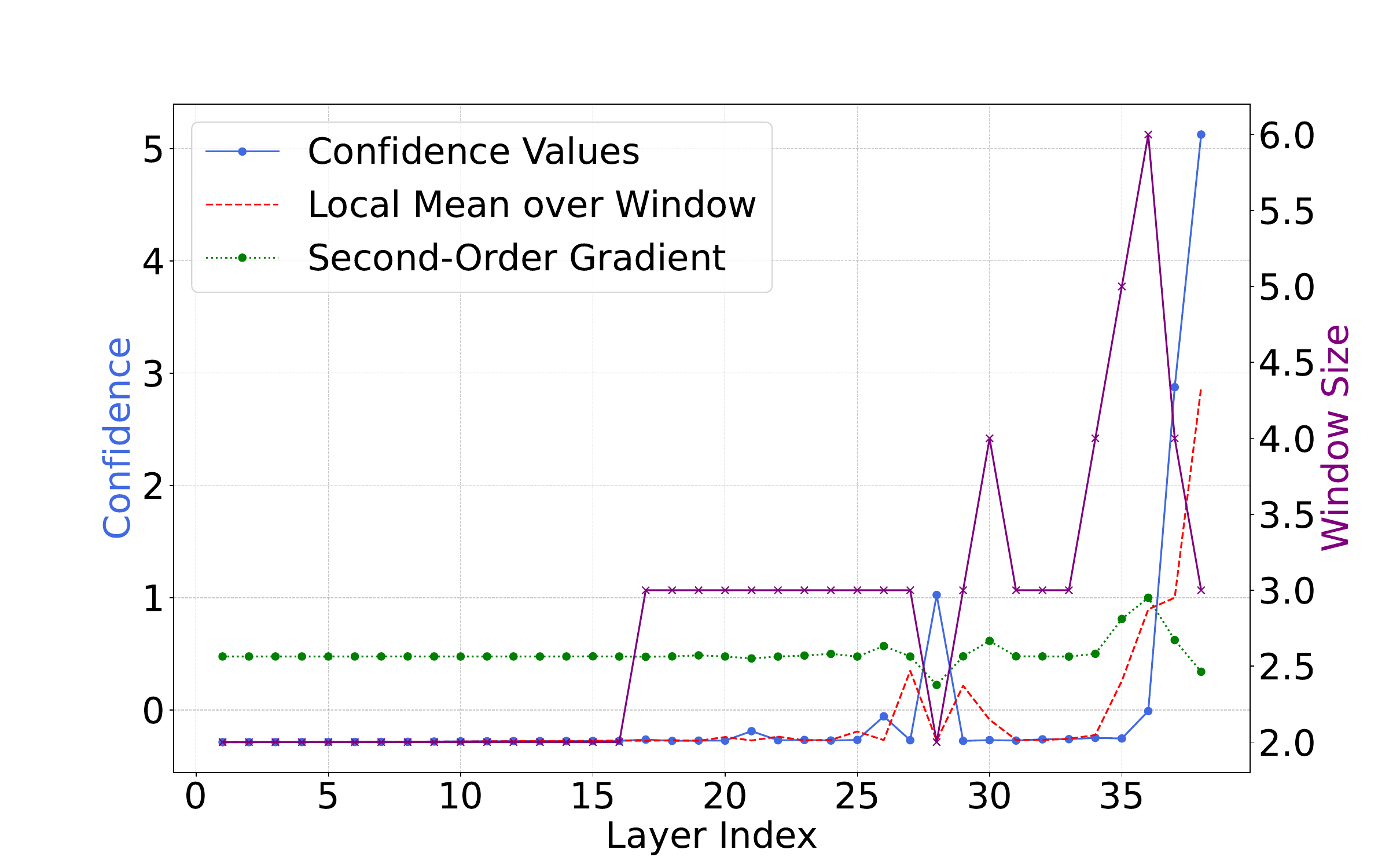}
    \caption{\centering Confidence, local mean, gradient values, and the window size for every layer with LLaMa-2-13B on GSM8K.}
    \label{fig1}
\end{figure}

We visualize the correlation between these different variables for one round of the layer set search process with the LLaMa-2-13B model over the CNN-DailyMail (CNN-DM) dataset in Figure \ref{fig1}. The plot shows the confidence values $c_i$ of each layer out of the 38 total considered layers (excluding the first and last layer) in blue and the quantified change in confidence values at every layer as the second-order gradient values in green. This change linearly affects the window size $w_i$ to compute the mean used as a threshold for layer filtering. This direct impact of the gradient on the window size can also be visualized in the plot in purple. We finally visualize the layer filtering threshold of every layer in the red plot. The confidence value of every layer will be compared to its corresponding layer threshold to decide if this layer would be included in the set forming the draft model.

\section{Detailed Speedup Comparison}
\label{app1}

Table \ref{tab5} presents the main results of our evaluation on the four LLaMa models. 
The results show that ConfLayers always achieves a speedup above 1 and up to 1.4$\times$, proving that our approach can accelerate LLM generation and provide an advantage over the vanilla approach. The table also shows that ConfLayers consistently achieves significantly higher speedup results than SWIFT, especially for larger models such as LLaMa-2-70B and LLaMa-3-70B. When these models are used for a summarization task over the CNN-DM dataset, our approach can yield a 1.371$\times$ and 1.378$\times$ speedup respectively compared to vanilla LLM generation while SWIFT can only reach a 1.299$\times$ and 1.258$\times$ speedup. Additionally, while SWIFT struggles to achieve speedup in cases like inference with LLaMa-2-13B on CNN-DM, ConfLayers is still able to accelerate inference by 1.162$\times$. This improvement is due to a key difference between ConfLayers and SWIFT, which is the objective function. ConfLayers prioritizes maximizing the number of accepted tokens by the target model instead of prioritizing optimizing for the acceptance rate as SWIFT does, since a draft model with high acceptance rate and low number of accepted tokens might not be the best performing. That is observed in the case where LLama-3-8B is used to run inference on Alpaca, as the acceptance rateof 0.0938$\times$ for ConfLayers is lower than the one SWIFT produces which is 0.965$\times$, but ConfLayers is still able to outperform SWIFT as the number of accepted tokens with ConfLayres is 4.383 which is significantly higher than the one with SWIFT being 3.414. An additional helping factor is the flexibility that ConfLayers provides during the search process. While SWIFT only allows a fixed skipping rate assigned to 45\% for smaller models and 50\% for larger ones, ConfLayers does not limit the search space to a certain number of layers to skip and expands the exploration set to any number of layers below or above a pre-defined threshold while conditioning the search on the performance of the draft model. Thus, the skip ratio for every model and task pair varies between 40 and 50\%.

\begin{table*}[h]
\caption{\centering Performance of ConfLayers compared to SWIFT and vanilla LLM generation with several LLaMa models over several tasks}
\label{tab5}
\begin{tabular}{cll|ll|ll|ll|ll}
\hline
\multicolumn{3}{c|}{\textbf{Model}}                                                                                                                                                       & \multicolumn{2}{l|}{\textbf{LLaMa-2-13B}} & \multicolumn{2}{l|}{\textbf{LLaMa-2-70B}} & \multicolumn{2}{l|}{\textbf{LLaMa-3-8B}} & \multicolumn{2}{l}{\textbf{LLaMa-3-70B}} \\ \hline
\multicolumn{3}{c|}{\textbf{Approach}}                                                                                                                                                    & \textbf{SWIFT}      & \textbf{CL}         & \textbf{SWIFT}      & \textbf{CL}         & \textbf{SWIFT}     & \textbf{CL}         & \textbf{SWIFT}      & \textbf{CL}        \\ \hline
\multicolumn{1}{c|}{\multirow{16}{*}{\textbf{Dataset}}} & \multicolumn{1}{l|}{\multirow{4}{*}{\textbf{CNN-DM}}}                                                   & \textbf{Speedup}      & 0.920$\times$              & \textbf{1.162$\times$}     & 1.299$\times$              & \textbf{1.371$\times$}     & 1.080$\times$            & \textbf{1.103$\times$}     & 1.258$\times$     & \textbf{1.378$\times$}             \\
\multicolumn{1}{c|}{}                                   & \multicolumn{1}{l|}{}                                                                                   & \textbf{$\alpha$}    & 0.940               & 0.972               & 0.962               & 0.958               & 0.950              & 0.941               & 0.952               & 0.972              \\
\multicolumn{1}{c|}{}                                   & \multicolumn{1}{l|}{}                                                                                   & \textbf{$\beta$}    & 45\%                & 42\%               & 50\%                 & 49\%               & 45\%               & 43\%                 & 50\%                 & 41\%              \\
\multicolumn{1}{c|}{}                                   & \multicolumn{1}{l|}{}                                                                                   & \textbf{$M$} & 2.566               & 4.333               & 3.384               & 3.114               & 3.523              & 3.344               & 3.974               & 5.129              \\ \cline{2-11} 
\multicolumn{1}{c|}{}                                   & \multicolumn{1}{l|}{\multirow{4}{*}{\textbf{GSM8K}}}                                                    & \textbf{Speedup}      & 1.085$\times$              & \textbf{1.125}$\times$     & 1.311$\times$              & \textbf{1.342$\times$}     & 1.087$\times$             & \textbf{1.096$\times$}     & 1.230$\times$              & \textbf{1.333$\times$}    \\
\multicolumn{1}{c|}{}                                   & \multicolumn{1}{l|}{}                                                                                   & \textbf{$\alpha$}    & 0.974               & 0.860               & 0.980               & 0.946               & 0.972              & 0.949               & 0.931               & 0.983              \\
\multicolumn{1}{c|}{}                                   & \multicolumn{1}{l|}{}                                                                                   & \textbf{$\beta$}    & 45\%                & 50\%               & 50\%                 & 45\%               & 45\%               & 43\%               & 50\%                 & 41\%              \\
\multicolumn{1}{c|}{}                                   & \multicolumn{1}{l|}{}                                                                                   & \textbf{$M$} & 3.403               & 3.393               & 3.997               & 3.620               & 2.970              & 3.080               & 3.321               & 4.442              \\ \cline{2-11} 
\multicolumn{1}{c|}{}                                   & \multicolumn{1}{l|}{\multirow{4}{*}{\textbf{\begin{tabular}[c]{@{}l@{}}WMT14 \\ (DE-EN)\end{tabular}}}} & \textbf{Speedup}      & 1.068$\times$              & \textbf{1.172$\times$}     & 1.105$\times$              & \textbf{1.351$\times$}     & 1.019$\times$             & \textbf{1.052$\times$}     & 1.281$\times$     & \textbf{1.346$\times$}             \\
\multicolumn{1}{c|}{}                                   & \multicolumn{1}{l|}{}                                                                                   & \textbf{$\alpha$}    & 0.961               & 0.954               & 0.943               & 0.994               & 0.944              & 0.955               & 0.978               & 0.988              \\
\multicolumn{1}{c|}{}                                   & \multicolumn{1}{l|}{}                                                                                   & \textbf{$\beta$}    & 45\%                & 45\%               & 50\%                 & 42\%               & 45\%               & 43\%               & 50\%                 & 42\%              \\
\multicolumn{1}{c|}{}                                   & \multicolumn{1}{l|}{}                                                                                   & \textbf{$M$} & 3.526               & 3.262               & 2.339               & 4.430               & 2.698              & 2.955               & 4.028               & 4.756              \\ \cline{2-11} 
\multicolumn{1}{c|}{}                                   & \multicolumn{1}{l|}{\multirow{4}{*}{\textbf{Alpaca}}}                                                   & \textbf{Speedup}      & 1.035$\times$     & \textbf{1.134$\times$}              & 1.228$\times$              & \textbf{1.353$\times$}     & 1.045$\times$             & \textbf{1.191$\times$}     & 1.055$\times$              & \textbf{1.156$\times$}    \\
\multicolumn{1}{c|}{}                                   & \multicolumn{1}{l|}{}                                                                                   & \textbf{$\alpha$}    & 0.944               & 0.967               & 0.966               & 0.985               & 0.965              & 0.938               & 0.917               & 0.946              \\
\multicolumn{1}{c|}{}                                   & \multicolumn{1}{l|}{}                                                                                   & \textbf{$\beta$}    & 45\%                & 42\%                 & 50\%                 & 45\%               & 45\%               & 47\%               & 50\%                 & 41\%              \\
\multicolumn{1}{c|}{}                                   & \multicolumn{1}{l|}{}                                                                                   & \textbf{$M$} & 2.318               & 3.643               & 2.597               & 4.228               & 3.414              & 4.383               & 3.206               & 2.746              \\ \hline
\multicolumn{3}{c|}{\textbf{Average Speedup}}                                   & 1.027$\times$ & \textbf{1.148$\times$} & 1.261$\times$ & \textbf{1.354$\times$} & 1.058$\times$ & \textbf{1.111$\times$} & 1.206$\times$ & \textbf{1.303$\times$}\\ \hline
\end{tabular}
\end{table*}

\section{Visualization of Skipped Layer Set Progression}
\label{app3}
\begin{figure}[h]
    \centering
    \includegraphics[width=\textwidth]{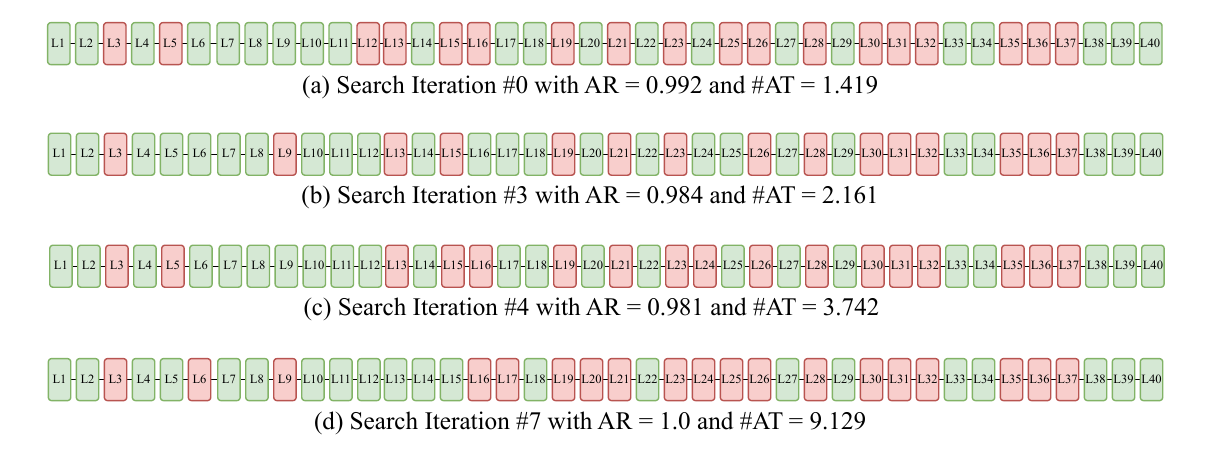}
    \caption{\centering Progression of the best skipped layer set (in red) with LLaMa-2-13B on GSM8K.}
    \label{fig10}
\end{figure}

We also visualize the best skip layer sets found in iterations \#0, \#3, \#4 and \#7 in Figure \ref{fig10}. The framework is initialized with uniform layer skipping which is used to generate the first set of tokens forming the context window. The tokens generated in this window will be used to extract the confidence of the layers and initiate the search for the first set of layers to form the draft model. The result of that search round is shown in Figure \ref{fig10} (a) with a mean number of accepted tokens of 1.42. The layer set is saved until better performing ones are found or until the search is halted. Seven search rounds later, the framework is able to find a layer set with a mean number of accepted tokens of 9.13. This shows that the framework is able to find draft models resulting in a high number of accepted tokens while conserving a relatively high acceptance rate. The final saved layer set is adopted for the remainder of the generation.

\end{document}